\documentclass{article}
\usepackage{ijcai24}

\usepackage{times}
\usepackage{soul}
\usepackage{url}
\usepackage[hidelinks]{hyperref}
\usepackage[utf8]{inputenc}
\usepackage[small]{caption}
\usepackage{graphicx}
\usepackage{amsmath}
\usepackage{amsthm}
\usepackage{booktabs}
\usepackage{algorithm}
\usepackage[switch]{lineno}
\usepackage{algpseudocode}

\usepackage{amsmath,amsfonts,bm}

\def\eqref#1{equation~\ref{#1}}

\def\1{\bm{1}}

\def\vx{{\bm{x}}}

\def\mI{{\bm{I}}}

\DeclareMathAlphabet{\mathsfit}{\encodingdefault}{\sfdefault}{m}{sl}
\SetMathAlphabet{\mathsfit}{bold}{\encodingdefault}{\sfdefault}{bx}{n}

\newcommand{\R}{\mathbb{R}}

\title{GLoD: Composing Global Contexts and Local Details in Image Generation}

\author{
    Moyuru Yamada
    \affiliations
    Fujitsu Limited
    \emails
    yamada.moyuru@fujitsu.com
}

\begin{document}

\maketitle

\begin{abstract}
Diffusion models have demonstrated their capability to synthesize high-quality and diverse images from textual prompts. However, simultaneous control over both global contexts (e.g., object layouts and interactions) and local details (e.g., colors and emotions) still remains a significant challenge. The models often fail to understand complex descriptions involving multiple objects and reflect specified visual attributes to wrong targets or ignore them. 
This paper presents Global-Local Diffusion (\textit{GLoD}), a novel framework which allows simultaneous control over the global contexts and the local details in text-to-image generation without requiring training or fine-tuning. It assigns multiple global and local prompts to corresponding layers and composes their noises to guide a denoising process using pre-trained diffusion models. Our framework enables complex global-local compositions, conditioning objects in the global prompt with the local prompts while preserving other unspecified identities.
Our quantitative and qualitative evaluations demonstrate that GLoD effectively generates complex images that adhere to both user-provided object interactions and object details. 
\end{abstract}

\begin{figure*}[t]
\begin{center}
    \includegraphics[width=0.92\linewidth, page=1]{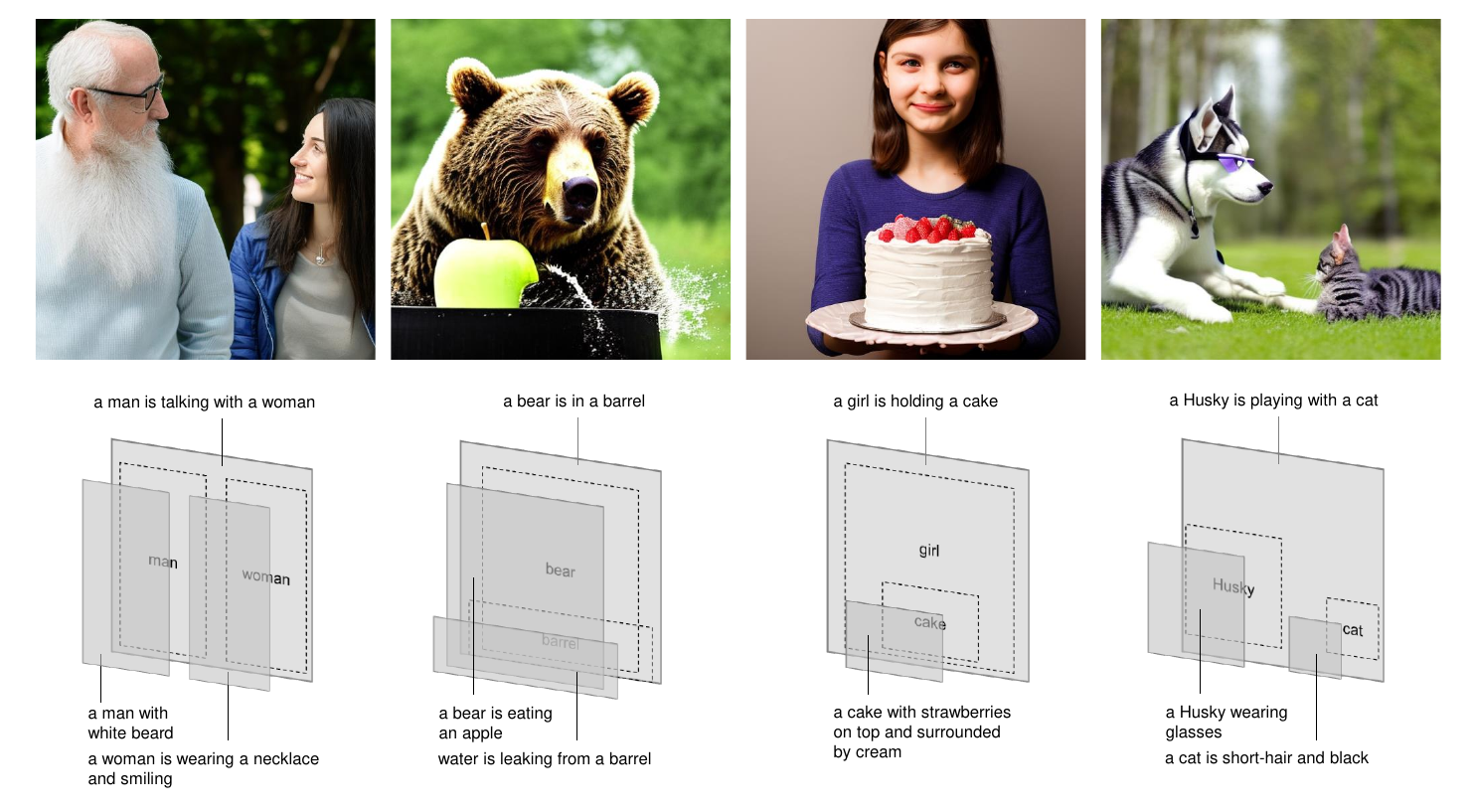}
\end{center}
\caption{Global-Local Diffusion (GLoD) takes multiple prompts as an input (e.g., a global prompt: 'a man is talking with a woman' and two local prompts: 'a man with white beard' and 'a woman is wearing a necklace and smiling') along with their layout and assigns noises obtained from them into corresponding layers with a pre-trained diffusion model. Then, the noises are effectively composed to generate an image. Details of objects in the global prompt are guided with the corresponding local prompts.}
\label{fig-main}
\end{figure*}

\section{Introduction}

Text-to-image generative models have emerged recently and demonstrated their amazing capabilities in synthesizing high-quality and diverse images from text prompts. Diffusion models ~\cite{dhariwal2021diffusion,ddpm,iddpm} are currently one of the state-of-the-art methods and widely used for the image generation. 
Despite their impressive advances in image generation, lack of control over the generated images is a crucial limitation in deploying them to real-world applications. 

To provide further controllability over diffusion models, researchers have put a lot of effort into control of object layouts, object interactions, and composition of objects. Training-free layout control ~\cite{trainingfree} takes a text prompt along with the object layout as an input and control the object position based on a loss between the input layout and attention maps. MultiDiffusion ~\cite{multidiffusion} places an object with specified details on a certain region using segmentation masks and a prompt for each segment. These methods work without requiring any additional training; however, they struggle to control both the global contexts (e.g., object interactions) and the local details (e.g., object colors and emotions) simultaneously. With a complex prompt containing multiple objects, the models often misinterpret specified local details, directing them to the wrong target or ignoring them, similar to the issues observed in Stable Diffusion ~\cite{latentdiffusion}. While splitting the complex prompt into multiple prompts allows the model to depict each object more accurately, handling the prompts independently poses limitations in addressing a global context that describes interactions and relationships between the multiple objects.

Another trial is training a model from scratch or finetuning a given diffusion model for better controllable generation with a task-specific annotation ~\cite{gligen,SpaText,Freestyle,zhang2023text2layer}. Some researchers have leveraged a scene graph as an input and introduced models which generate images from the scene graph to control the object interactions in the generated images ~\cite{farshad2023scenegenie} and ~\cite{yang2022diffusionbased}. These methods have shown superior results, while they often require a high computation cost and a long development period. This makes it difficult to leverage various pre-trained diffusion models.



This paper proposes Global-Local Diffusion (\textit{GLoD}), a novel diffusion framework that controls both global contexts and local details simultaneously using a pre-traind diffusion model without requiring any additional training or finetuning. 
GLoD takes as input global prompts that describe entire image including object interactions, and local prompts that specify object details along with their position in the form of a bounding box. The diffusion model predicts noises from the prompts separetely and then the noises are composed to guide the denoising process. 
For example, instead of giving a complex prompt such as ’a man with white beard is talking with a smiling woman wearing a necklace', we decompose it into multiple prompts, a global context 'a man is talking with a woman' and two local details 'a man with white beard' and 'a woman is wearing a necklace and smiling'. The details of the man and the woman in the global prompt are guided with each local prompt in the image generation process. GLoD also controls the object layout with the given bounding boxes. The generated examples are as shown in Fig. \ref{fig-main}.




Our framework enables both global-global compositions and global-local compositions. The global-global composition composes foreground and background similar to the existing method ~\cite{compositional}, while the global-local composition composes the global context and the object details. Since a local prompt can be a global prompt for other local prompts, GLoD allows us to compose more than two layers. 
In addition, unlike the existing methods that may changes object identities even by just adding a single attribute, GLoD only changes the object details specified by the corresponding local prompts while preserving other identities. This feature enables users to control the generated image interactively.

In order to assess the effectiveness of our proposed method quantitatively, we build a new test set to evaluate the controllability over the global contexts and local details in the image generation rather than the image quality.

Our key contributions are summarized as follows\footnote{The code will be made publicly available upon acceptance.}: 
\begin{itemize}
    \item 
We propose Global-Local Diffusion (GLoD), a simple and yet effective framework for diffusion-based image synthesis and editing which enables controlling both global contexts and local details simultaneously. 

    \item Through quantitative and qualitative evaluations, we demonstrate that our proposed method can effectively generate complex images by composing the multi-prompts describing object interactions and the details.
\end{itemize}

\begin{figure*}[t]
\begin{center}
    \includegraphics[width=0.8\linewidth, page=1]{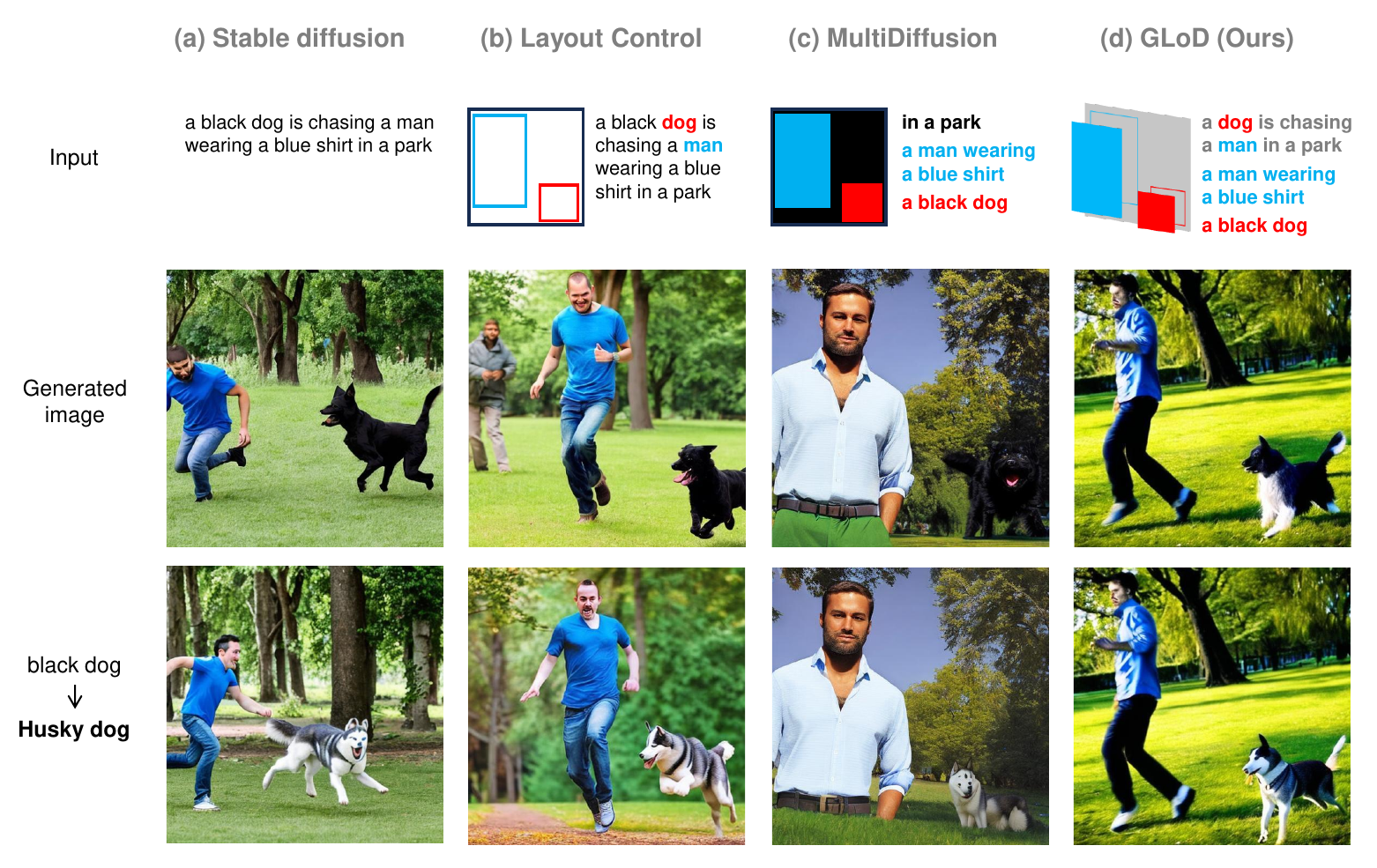}
\end{center}
\caption{GLoD enables controlling global contexts (interaction between a dog and a man, their layouts) and local details (the dog is black, the man is wearing a blue shirt) independently. Local details can be specified (black dog $\rightarrow$ Husky dog) while preserving the global contexts. Note that this is not image editing. We generate images from the text prompts and the layout.}
\label{fig-related}
\end{figure*}

\section{Related Work}
\subsection{Diffusion Models}
Diffusion models ~\cite{dhariwal2021diffusion,ddpm,iddpm} has attracted a lot of attention as a promising class of generative models that formulates the data generation process as an iterative denoising procedure. The models take a Gaussian noise input $\vx_T \sim  \mathcal{N} (\textbf{0},\mI)$ and transform it into a sample $\vx_0$ through the series of gradual denoising steps $T$. The sample should be distributed according to a data distribution $q$. Many research works focus on improving the diffusion process to speed up the sampling process while maintaining high sample quality ~\cite{iddpm,Karras2022edm}. 
The latent diffusion model ~\cite{latentdiffusion} has also been developed to address this issue and applied the diffusion process in latent space instead of pixel space to enable an efficient sampling.
While the diffusion models have originally shown great performance in image generation, enabling effective image editing and image
inpainting ~\cite{meng2022sdedit,blend}, these models been successfully used in various domains, including video ~\cite{imagenvideo}, audio ~\cite{wavegrad}, 3D scenes ~\cite{diffrf}, and motion sequences ~\cite{tevet2023human}. Although this paper focuses on image generation, our proposed framework may further be applied in such domains.

\subsection{Controllable Image Generation with Diffusion}
Diffusion models are first applied to text-to-image generative models, which generate an image conditioned on a free-form text description as an input prompt. Classifier-free guidance ~\cite{classifierfree} plays important role in conditioning the generated images to the input prompt. Recent text-to-image diffusion models such as DALL-E 2 ~\cite{dalle2}, Imagen ~\cite{imagen}, and Stable Diffusion ~\cite{latentdiffusion} has shown remarkable capabilities in image generation.
On the other hand, recent studies ~\cite{trainingfree,multidiffusion,layoutdiff,localizing} have stressed the inherent difficulty in controlling generated images with a text description, especially in the control over (i) object layout and (ii) visual attributes of objects.
To gain more control over the object layout, some works have leveraged bounding boxes or segmentation masks as an additional input along with text prompts.
Training-free layout control ~\cite{trainingfree} takes a single prompt along with a layout of objects appeared in the prompt as shown in Fig. \ref{fig-related} (b). Object layout is given in a form of the bounding. Layout control extracts attention maps from a pre-trained diffusion model and updates the latent embeddings of the image based on an error between the input bounding boxes and the attention maps. Since this method simply uses the pre-trained diffusion model to generate images, it inherits the difficulties in control over the visual attributes of objects, i.e., the local details. Also, the generated image may largely change even if a single word is added or replaced in the prompt due to the fact that the input prompt describes both the global contexts and the local details.
MultiDiffusion ~\cite{multidiffusion} and SceneComposer ~\cite{sceneComposer} take multiple prompts along with their corresponding segmentation masks as a region as shown in Fig. \ref{fig-related} (c). They effectively control the object layout and visual attributes of each object. However, they cannot handle a prompt describing interactions between those objects, i.e., the global contexts. Basically, they just place a specific object described by the prompt in a certain region. Thus, if the input prompt is replaced (black dog $\rightarrow$ Husky dog), the new object (Huskey dog) does not inherit the contexts (e.g., posture) from the replaced object (black dog).
Unlike these methods, our method aims to control both the global contexts and the local details simultaneously. Since we treat the global contexts and the local details separately, the global contexts are preserved even if the local details are changed, as shown in Fig. \ref{fig-related} (d). Some studies ~\cite{farshad2023scenegenie,yang2022diffusionbased} focused on image synthesis from scene graphs for better control over the complex relations between multiple objects in the generated images. However, these works require costly
extensive training on curated datasets. They regard a complex scene graph as an input prompt, while our approach decomposes the complex prompts into multiple simple prompts and does not require any training or finetuning.






\begin{figure*}[t]
\begin{center}
    \includegraphics[width=0.75\linewidth, page=1]{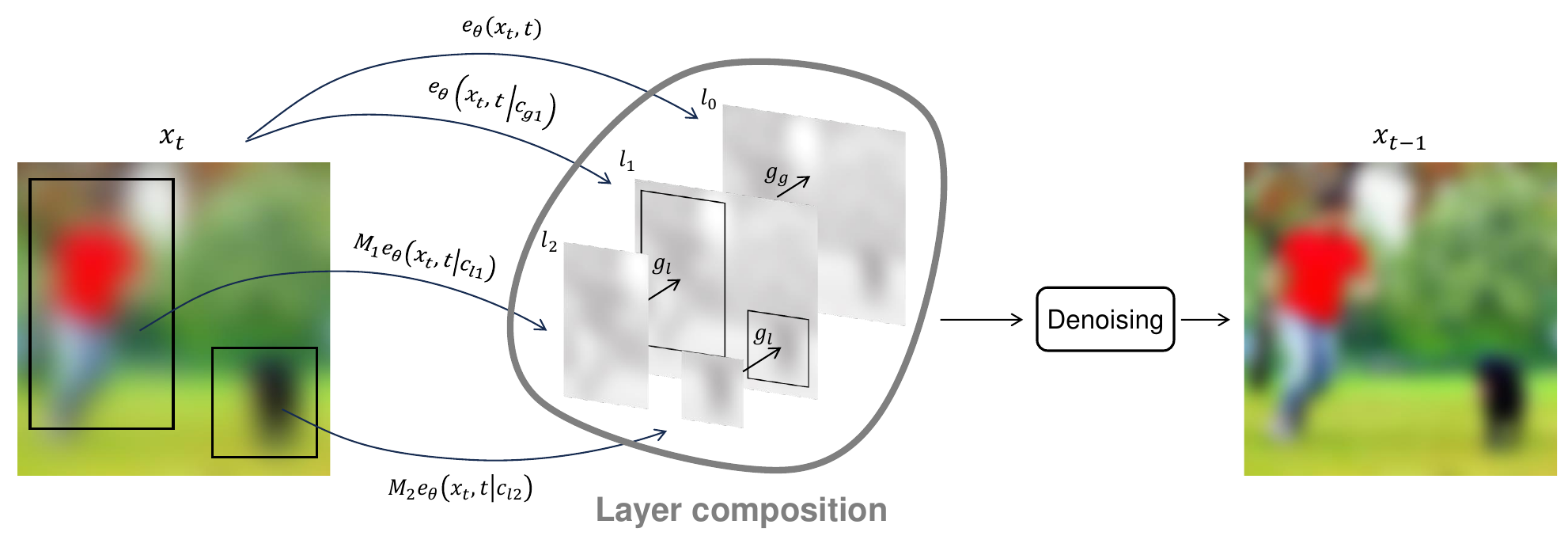}
\end{center}
\caption{GLoD composes multiple layers. Unconditional noise and noises conditioned on global contexts (e.g., interactions) or local details (e.g., color) are assigned to separate layers ($l_0$, $l_1$, $l_2$). Those layers are then composed with global guidance $g_g$ and local guidance $g_l$.}
\label{fig-layers}
\end{figure*}

\subsection{Layered Image Generation and Editing}
Some recent works ~\cite{zhang2023text2layer,li2023layerdiffusion,liao2023textdriven} have proposed layered image generation and editing. They considers two layers, foreground and background, and enables to control them individually with a segmentation mask of the foreground object. Their models are needed to be trained with proposed losses. Unlike them, our goal is to control the global contexts and the local details simultaneously without requiring the training and the accurate segmentation masks as shown in Fig. \ref{fig-main}. Our framework enables to control local details while keeping the global contexts by composing the multiple layers, where the global layer may represent the foreground or background and the local layer may represent the details of the objects in the global layer. The global layer and the local layer are not independent but have a whole-part relationship. Our framework can also handle more than two layers as shown in Fig. \ref{fig-editing}.

\subsection{Compositional Generation}
The compositional generation is an approach to generate the complex images by composing a set of diffusion models, with each of them modeling a certain component of the image. This approach has been an essential direction for image-to-text models because it is difficult for the current models to handle complex prompts where multiple concepts are squeezed. Recently, ~\cite{compositional} has demonstrated successful composition of independent concepts (e.g., “a bear” and “in a forest”) by adding estimated score for each concept. ~\cite{trainingfreestr} has also proposed another approach which can be directly merged into the cross-attention layers. Inspired by the first approach, we propose a novel method to compose whole-part concepts (e.g., "a bear is eating an apple" and "the apple is green").








\section{Method}
Our goal is to generate images where given global contexts and local details are reflected. In this section, we introduce Global-Local Diffusion (GLoD) to compose the global context and the local detail with pre-trained diffusion models.

\subsection{Compositions of Diffusion Models}

We consider a pre-trained diffusion model, which takes a text prompt $y \in \mathcal{Y}$ as a condition and generates a intermediate image $\vx_t \in \mathcal{I} = \R^{H{\times}W{\times}C}$:

\begin{equation}
\vx_{t-1}=\Phi(\vx_t|y).
\end{equation}

The diffusion models are also regarded as Denoising Diffusion Probabilistic Models (DDPMs) where generation is modeled as a denoising process. The objective of this model is to remove a noise gradually by predicting the noise at a timestep $t$ given a noisy image $x_t$. To generate a less noisy image, we sample $x_{t-1}$ until it becomes realistic over multiple iterations:

\begin{equation}
\vx_{t-1}=\vx_{t}-\epsilon_{\theta}(\vx_t, t) + \mathcal{N} ( 0 , {\sigma_t}^{2}I),
\end{equation}
where $\epsilon_{\theta}(\vx_t, t)$ is the denoising network. ~\cite{compositional} has revealed that the denoising network or score function can be expressed as a compositions of multiple score functions corresponding to an individual condition $c_i$.

\begin{equation}
\hat{\epsilon}(\vx_t, t) = \epsilon_{\theta}(\vx_t, t)+\sum_{i = 0}^{n}{w_{i}(\epsilon_{\theta}(\vx_t, t|c_i)-\epsilon_{\theta}(\vx_t, t))}.
\label{comp}
\end{equation}
where $\epsilon_{\theta}(\vx_t, t|c_i)$ predicts a noise conditioned on $c_i$ and $\epsilon_{\theta}(\vx_t, t)$ outputs an unconditional noise.
This equation only focuses on composing individual conditions over entire image, e.g., a foreground condition like 'a boat at the sea' and a background condition like 'a pink sky'.

We regard $\epsilon_{\theta}(\vx_t, t|c_i)-\epsilon_{\theta}(\vx_t, t)$ as a guidance $g_i$, which guides the unconditional noise toward the noise conditioned on a given condition $c_i$. Then, the composed denoising network is viewed as a composition of the guidance.
\begin{equation}
\hat{\epsilon}(\vx_t, t) = \epsilon_{\theta}(\vx_t, t)+\sum_{i = 0}^{n}{w_{i}g_i}.
\end{equation}

\begin{figure}[!t]
\begin{algorithm}[H]
    \caption{GLoD sampling.}
    \label{alg1}
    \begin{algorithmic}[1]
    \Require{Diffusion model $\epsilon_{\theta}(\vx_t, t)$, global scales $w_{i}$, local scales $w_{j}$, global conditions $c_{gi}$, local conditions $c_{lj}$}, object region masks $M_{j}$
    \State{Initialize sample $\vx_t \sim  \mathcal{N} (\textbf{0},\mI)$}
    \For {$t=T,\ldots,1$}
        \State{$\vx_t \leftarrow f(\vx_t, c_{gi}, M)$} \Comment{apply layout control $f$}
        \State{$\epsilon_i \leftarrow \epsilon_{\theta}(\vx_t, t|c_{gi})$} \Comment{scores for global condition $c_{gi}$}
        \State{$\epsilon_j \leftarrow \epsilon_{\theta}(\vx_t, t|c_{lj})$} \Comment{scores for local condition $c_{lj}$}
        \State{$\epsilon \leftarrow \epsilon_{\theta}(\vx_t, t)$\Comment{unconditional score}} 
        \State{$\epsilon_b \leftarrow \epsilon_i, \epsilon_j$ \Comment{Assign $\epsilon_i$ and $\epsilon_j$ to $\epsilon_b$}} 
        \State{$g_g \leftarrow \sum_{i = 0}^{k}{w_{i}(\epsilon_i-\epsilon)}.$}\Comment{global guidance-Eq. \ref{eq-global}}
        \State{$g_l \leftarrow \sum_{j = 0}^{m}{w_{j}M_{j}(\epsilon_j-\epsilon_{b})}.$}\Comment{local guidance-Eq. \ref{eq-local}}
        \State{$\vx_{t-1} \sim \mathcal{N} (\vx_{t}-(\epsilon+g_g+g_l), {\sigma_t}^{2}I$) \Comment{sampling}}
    \EndFor
    \end{algorithmic}
\end{algorithm}
\end{figure}

\subsection{Layer Composition}
We propose GLoD to extend the above concept to a composition of a global condition and a local conditions, i.e., interactions between objects and object details.
We consider a set of global conditions $c_{g}=(c_{g1}, ..., c_{gk})$ and a set of local conditions $c_{l}=(c_{l1}, ..., c_{lm})$. 
We also introduce a diffusion layer $l=(l_0,...,l_t)$, where each layer contains one or more noises derived from the corresponding prompt as shown in Fig. \ref{fig-layers}. 
For example, with given a global prompt ($c_{g1}$) and two local prompts ($c_{l1}$ and $c_{l2}$), an unconditional noise can be assigned on a layer $l_0$, a layer $l_1$ contains a noise derived from the global prompt, and layer $l_2$ contains noises obtained from the local prompts. 
We compose the assigned noises with two ways of guidance: (i) \textit{global guidance}, which guides the image with global conditions by the following equation:
\begin{equation}
g_{g}=\epsilon_{\theta}(\vx_t, t|c_{g})-\epsilon_{\theta}(\vx_t, t),
\label{eq-global}
\end{equation}
where the unconditional noise is always a base noise $\epsilon_b$. This is also well known as Classifier-free guidance ~\cite{classifierfree}. With two global conditions, their global guidance is summed as a global-global composition. The classifier-free guidance works well on the global-global compositions, while it does not work effectively when we compose the global condition and the local condition since their conditions have some overlap. Thus, we newly propose (ii) \textit{local guidance}, which guides an object on the base layer $b$ conditioned on a condition $c_{b}$ with a local condition $c_{j}$.

\begin{equation}
g_l=M_{j}(\epsilon_{\theta}(\vx_t, t|c_{j})-\epsilon_{\theta}(\vx_t, t|c_{b})),
\label{eq-local}
\end{equation}
where $M_{j}\subset\{0,1\}^{H{\times}W}$ is a region mask of $j$-th region corresponding to the condition $c_{j}$. 
In Fig. \ref{fig-layers}, two local guidance is added to the global guidance as a global-local composition.
The intuition behind the local guidance is that an image region guided with a word 'dog' in a global prompt can be regarded as an unconditional 'dog' and guided with a local prompt by emphasizing the difference between the global and local conditions. In the end, decomposed global prompts and local prompts are effectively composed by our proposed guidance.


\begin{figure*}[!t]
\begin{center}
    \includegraphics[width=0.9\linewidth, page=1]{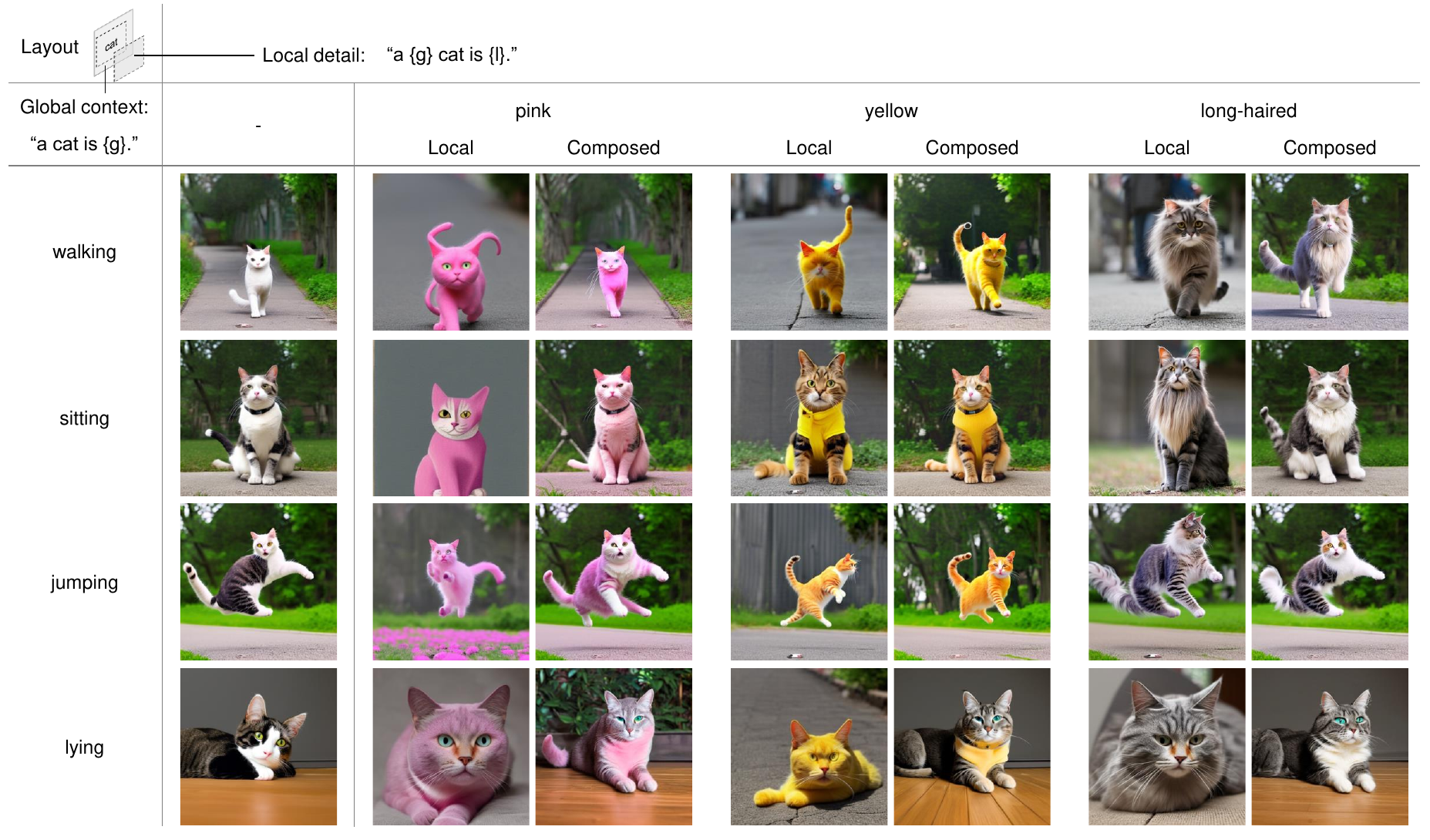}
\end{center}
\caption{GLoD for a single object. The images in the first column and 'Local' columns are sampled only from the global context (global images) and the local detail (local images) as an input prompt, respectively. The images in 'Composed' columns are sampled using our method, which effectively applies local detail (e.g., long-haired) to the object in the image while preserving the global contexts (i.e., object layouts and object postures). }
\label{fig-glocal-single}
\end{figure*}

\begin{figure}[!t]
\begin{center}
    \includegraphics[width=\linewidth, page=1]{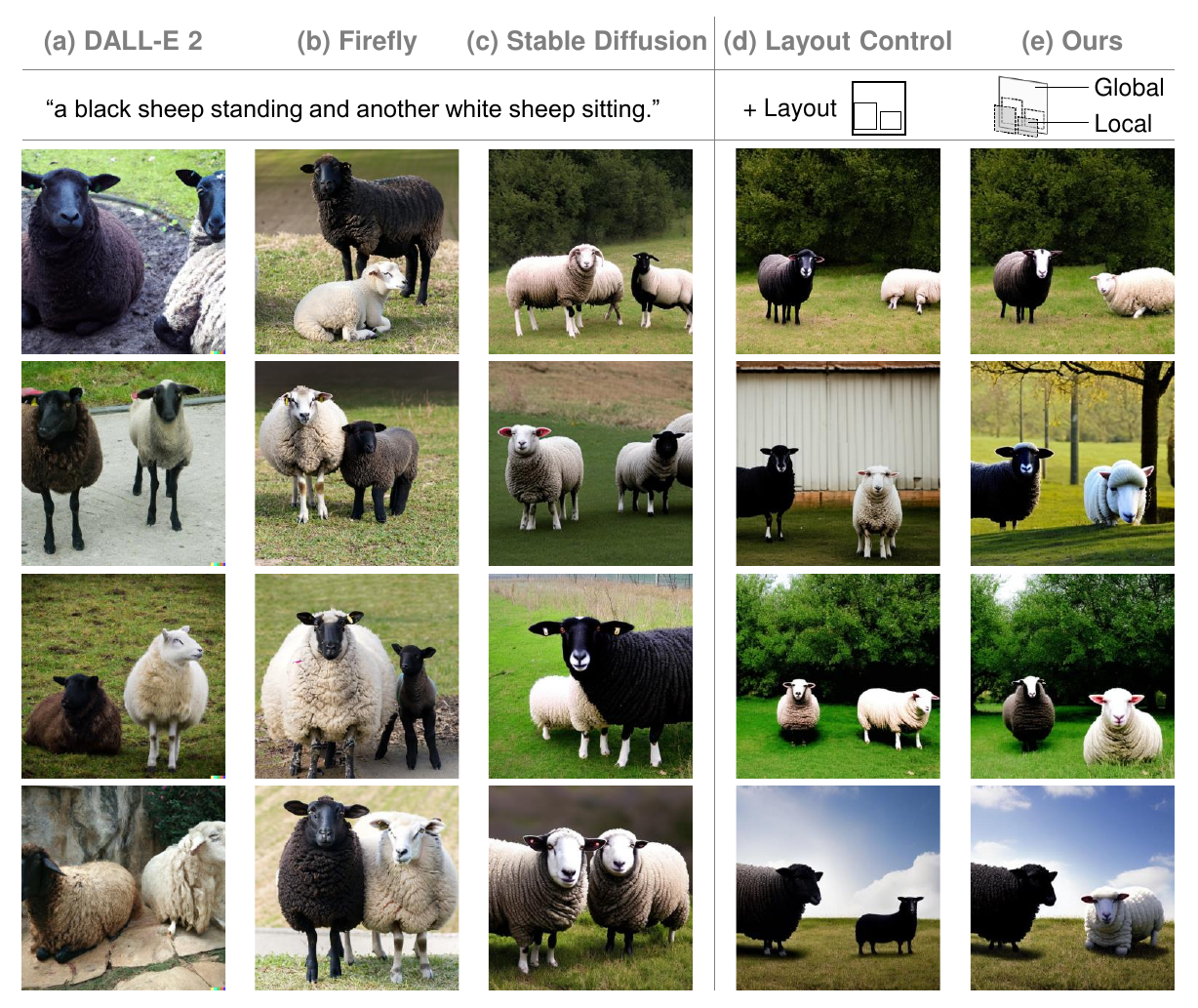}
\end{center}
\caption{GLoD for multiple objects. Our method (e) can control attributes of each sheep, while the other methods fail to reflect the specified attributes to the correct targets.}
\label{fig-glocal-multi}
\end{figure}

\subsection{Layout Control}
\label{method-layout-control}
Without any layout control, objects described in a given prompt appear somewhere in a generated image. To effectively compose the global noise and the local noise, we use Training-free layout control ~\cite{trainingfree}. More specifically, we use the backward guidance to control the layout of the objects in their layer before computing the global noise. 

Algorithm \ref{alg1} provides the pseudo-code for composing diffusion noises with GLoD. Our method composes noises obtained with pre-trained diffusion models during inference without any additional training or finetuning.

\section{Results}

\subsection{Evaluation Metrics}
We build a new test set to evaluate the controllability over the global contexts and local details in the image generation rather than the image quality. The test set contains 2500 samples where each sample contains a full text (e.g., 'a beard man is talking to a woman with earrings.'), a global text (i.e., 'a man is talking to a woman.'), local texts for a subject and an object ('a beard man.' and 'a woman with earrings.'), and a layout of the subject and the object. This test set design allows us to compute an alignment score using CLIP similarity ~\cite{clip}. We compute a global alignment score $S_g$ from an entire image and the global text, and similarly a local alignment score for the subject $S_{ls}$ and the object $S_{lo}$ from a region of them in the image and the corresponding local text. We also introduce an infection score $S_i$ that indicates how much undesirable effects the subject and object have. This can be an important metric since a prompt for an object A may also unintentionally affect another object B. We compute the similarity between a region of A and a prompt for B, similarly for the region of B and the prompt for A, and average them. See more details in Appendix \ref{appendix:testset}.

\subsection{Implementation Details}

We evaluate our method on the following conditions. In all experiments, we used Stable
Diffusion ~\cite{latentdiffusion} as our diffusion model, where the diffusion
process is defined over a latent space $I = R^{64\times64\times4}$, and
a decoder is trained to reconstruct natural images in higher
resolution $[0, 1]^{512\times512\times3}$. We use the public implementation
of Stable Diffusion by HuggingFace, specifically the Stable Diffusion v2.1 trained on the LAION-5B dataset ~\cite{schuhmann2022laionb} as the pre-trained image generation model. We also set Euler Discrete Scheduler ~\cite{Karras2022edm} as the noise scheduler. As our layout control (see \ref{method-layout-control}) we use the backward guidance ~\cite{trainingfree}. All the experiments are running on one A30 GPU.

We compare our GLoD with other state-of-the-art training-free methods, including Training-free layout control ~\cite{trainingfree} and MultiDiffusion ~\cite{multidiffusion}, and strong baselines, including OpenAI DALL-E 2 ~\cite{dalle2} and Adobe Firefly ~\cite{firefly}. We use the publicly available official codes and websites, and follow their instructions. 

\subsection{Image Generation with GLoD}

\noindent
{\bf GLoD for a single object.} We first demonstrate global-local compositions for a single object using GLoD for a better understanding, while our main targets are more complex scenes including multiple objects as shown in Fig. \ref{fig-main}. In Fig. \ref{fig-glocal-single}, our method generates diverse samples which comply with compositions of a global context (e.g., a cat is walking) and a local detail (e.g., a walking cat is pink). The images in the first column and 'Local' columns are sampled only from the global context (global images) and the local detail (local images) as an input prompt, respectively. The images in 'Composed' columns are sampled using our method, where the goal is to apply local detail (i.e., specified visual attribute) to the object in the image while preserving the global contexts. Although we generated all the images with the same seed, the posture of the cat is largely different in the corresponding global image and local image (e.g., the image of 'a cat is sitting' vs the image of 'a sitting cat is pink'). Our method can effectively generate the images from the global context and the local detail along with the layout (see 'Composed'). Note that this is image generation not the editing. The generated image retains most of the global contexts, including postures and head directions. In a few cases, the visual attribute of the object changes only partially (e.g., composition of 'a cat is lying' and 'a lying cat is yellow').

{\bf GLoD for multiple objects.} We then compare GLoD with the other state-of-the-art methods in generating more complex scene including multiple objects. In Fig. \ref{fig-glocal-multi}, we try to generate images of a complex scene where there are multiple objects in the same category (sheep in this case) and each of them has different attributes (a sheep is black and standing, another sheep is white and sitting). We show first four images generated by official web application of (a) DALL-E 2 and (b) Firefly at first and second column, respectively. They can generate high-quality images, but they often fail to reflect the specified attributes to the correct targets. We then find four seeds which generate failure samples of Training-free layout control and generate images using Stable diffusion and our method with those seeds. Both the layout control and our method use the same object layout as an additional input. We set 'a black sheep and another white sheep' as a global context, 'a sheep is black and standing' as a local detail of a sheep, and 'a sheep is white and sitting' as a local detail of the another sheep. We make the global context similar to the original prompt to compare the generated images easily. Figure \ref{fig-glocal-multi} shows that our method effectively controls the attributes of each object in the image.

{\bf Quantitative evaluation.} Table \ref{tab:scores} shows quantitative evaluation of controllability over the global context and the local details. Our method improves the local alignment scores $S_{ls}$ and $S_{lo}$ while keeping the global alignment score $S_g$ almost the same. Compared to Multi Diffusion, the proposed method shows a superior overall alignment score $S_{gl}$. Note that Multi Diffusion cannot handle the global prompt describing the object interaction, and thus shows significantly lower global alignment score.  The local alignment score for an object $S_{lo}$ lags considerably behind that for a subject $S_{ls}$.
We found that the subjects (e.g., woman) are often turning their backs on as a result of complying with the given global prompt (e.g., a man is talking with a woman). Therefore, the alignment score $S_{lo}$ becomes low because some of specified attributes are not visible in such cases. GLoD also improves the infection score $S_{i}$ against the baselines. This result indicates that our method reduces the mis-alignment between the prompt and the generated image.

{\bf GLoD for complex scenes.} Figure \ref{fig-main} shows other samples depicting more complex scenes, where we give an interaction between the objects as a global context and also specify the local details. Instead of giving a complex prompt such as 'a Husky wearing glasses is playing with a black short-hair cat', we decompose and handle them separately (i.e., 'a Husky is playing with a cat', 'a Husky wearing glasses', and 'a cat is short-hair and black') to effectively synthesis complex visual scenes.

\begin{table}
    \centering
    \begin{tabular}{lccccc}
        \toprule
        Methods  & $S_g$ $\uparrow$& $S_{ls}$ $\uparrow$& $S_{lo}$ $\uparrow$& $S_{gl}$ $\uparrow$& $S_i$ $\downarrow$ \\
        \midrule
        Stable Diffusion     & 24.2          & $-$ & $-$ & $-$ & $-$        \\
        Layout control  & 24.4          & 21.2 & 20.5 & 22.6 & 15.6     \\
        GLoD (ours) &  \textbf{24.6}          & \textbf{23.3} & \textbf{20.9} & \textbf{23.3} & \textbf{14.5}       \\
        \midrule
        MultiDiffusion          & 21.2          & 24.5 & 24.2 & 22.7 & 14.9        \\
        \bottomrule
    \end{tabular}
    \caption{Evaluation of controllability over the global context ($S_g$) and the local details for a subject and an object ($S_{ls}$ and $S_{lo}$). $S_{gl}$ represents an average of the global alignment score $S_g$ and the local alignment scores $S_{l\{s,o\}}$. $S_i$ denotes an infection score that indicates how much undesirable effects the subject and the object have.}
    \label{tab:scores}
\end{table}

\subsection{Global-Local Composition}

\begin{figure}[t]
\begin{center}
    \includegraphics[width=\linewidth, page=1]{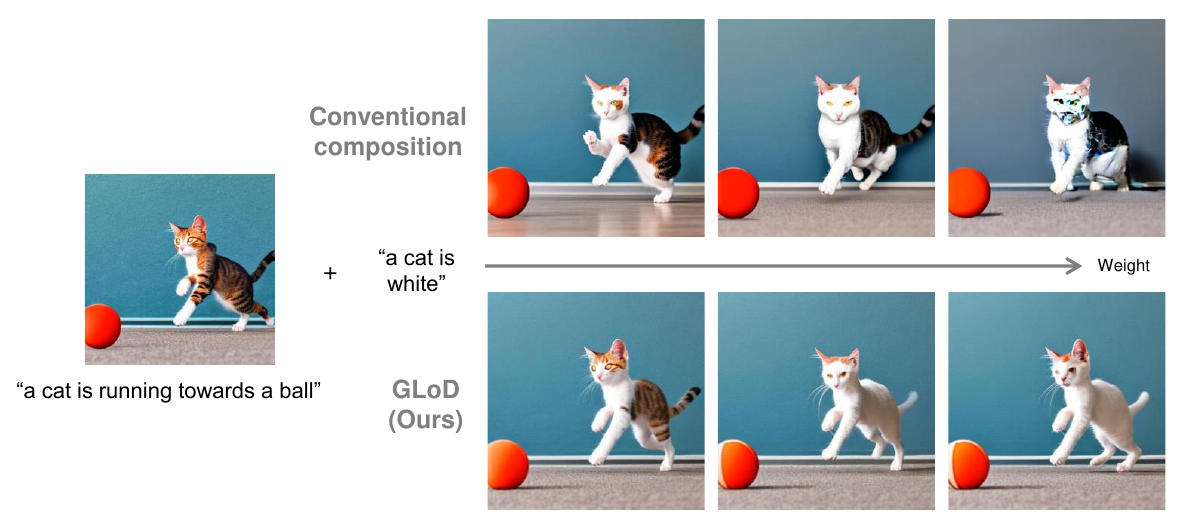}
\end{center}
\caption{Comparison between our global-local composition (bottom) and the conventional composition (top). Our method can change the detail of the object while preserving the global context by composing two prompts, whereas the conventional method often fails because they regard the prompts as two independent concepts. }
\label{fig-composition}
\end{figure}

We compare our global-local composition with a conventional composition ~\cite{compositional}. Since the conventional composition aims to compose independent concepts (e.g., foreground and background) by adding estimated score for each concept, it often fails to compose overlapped concepts (e.g., 'running cat' and 'white cat') as shown in Fig. \ref{fig-composition} (top). Our layer composition can effectively compose such overlapped concepts as shown in Fig. \ref{fig-composition} (bottom).





\subsection{Image Editing with GLoD}

\begin{figure}[t]
\begin{center}
    \includegraphics[width=\linewidth, page=1]{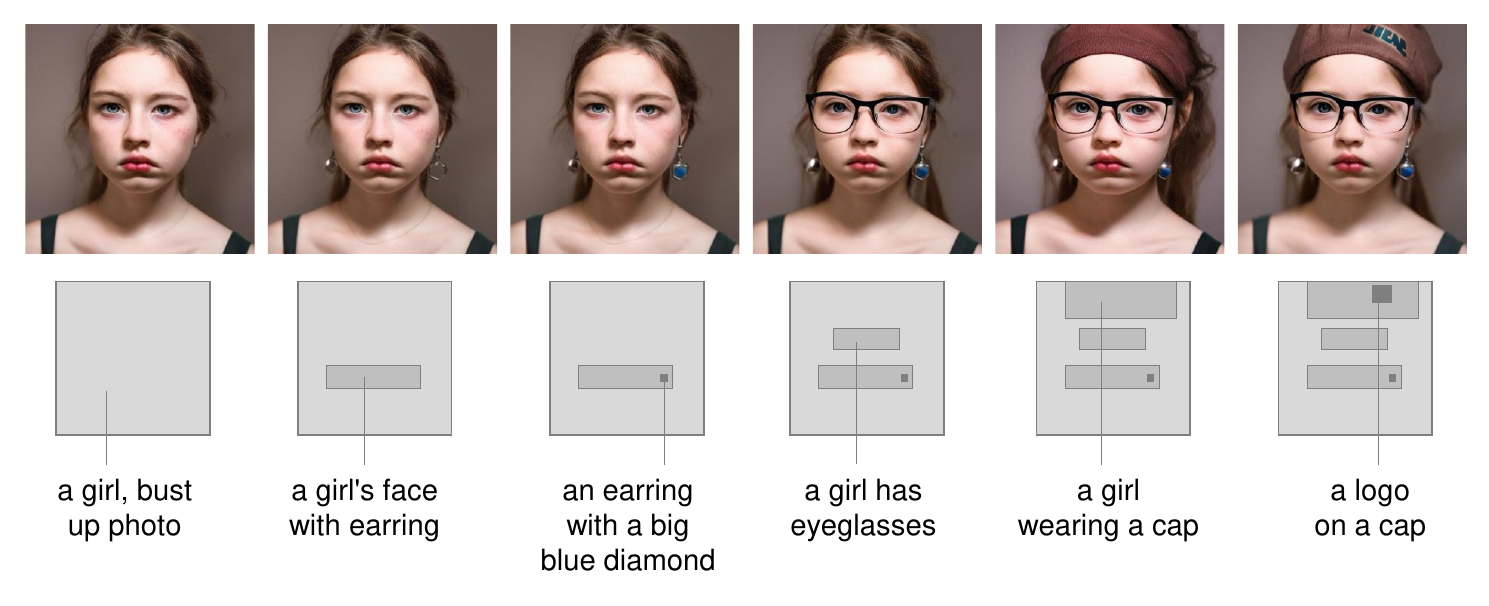}
\end{center}
\caption{GLoD also enables layered image editing, where new objects can be added on a certain region using additional prompts. Final image (right end) can be generated in one inference by composing six prompts. }
\label{fig-editing}
\end{figure}

Figure \ref{fig-editing} shows edited image samples with GLoD. GLoD enables layered image editing, where new objects can be added on a certain region using additional prompts. Details of the objects on a base layer (e.g., earring) can be guided with the prompts on the upper layer (e.g., adding a diamond) with our layer composition. Final image (right end) can be generated in one inference by composing multiple prompts (six in this case).

\section{Conclusion}
Image generation with simultaneous control over global contexts and local details is still an open challenge. We proposed GLoD, a simple and yet effective framework which composes a global prompt describing an entire image and local prompts specifying object details with a pre-trained diffusion model. Our framework can handle both global-global compositions and global-local compositions without requiring any additional training or finetuning. Through the qualitative and quantitative evaluations, we demonstrated that GLoD effectively generates images that include interactions between objects with detailed visual control, improving the alignment scores and reducing the undesirable affects.
A limitation we found is that the object appearance may change only partially when the latent of the object is significantly different between the global and the local.







\clearpage
\newpage

\bibliographystyle{named}
\bibliography{ijcai24}

\clearpage
\newpage

\appendix

\begin{figure}[!ht]
\begin{center}
    \includegraphics[width=0.9\linewidth, page=1]{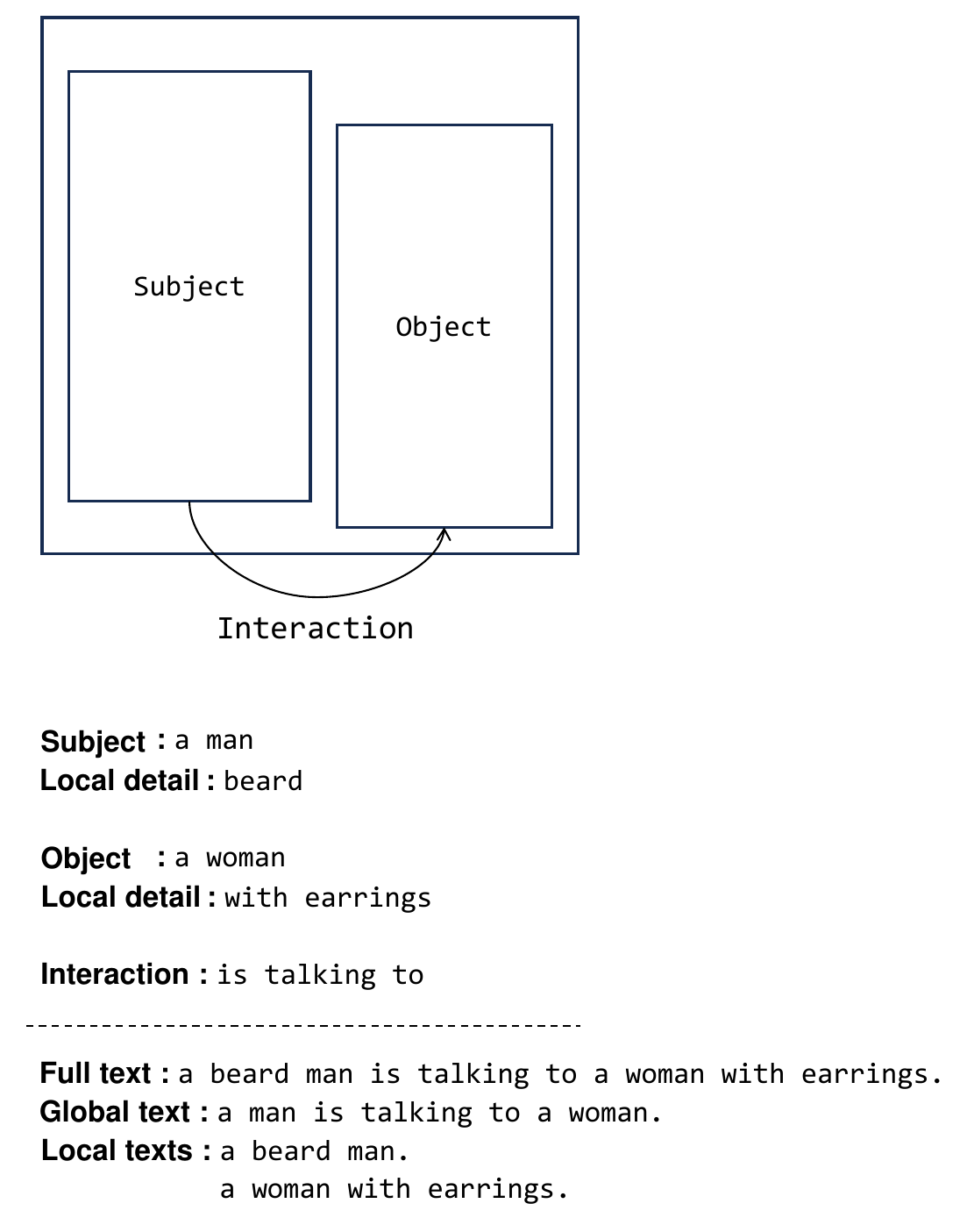}
\end{center}
\caption{An example in our test set. We give a subject and an object with their location, their attributes, and an interaction between them. }
\label{fig-appendix-testset}
\end{figure}

\section{Test set}
\label{appendix:testset}

We generated 2500 samples for testing the controllability over the global contexts and local details in the image generation. Fig. \ref{fig-appendix-testset} shows an example in our test set. Each sample contains a full text, a global text, and a local text for a subject and an object along with their layout. The samples are automatically generated based on a template, where we give a subject text, an object text, their details, a layout, and an interaction between them as an input. Since the object layout control is not our main focus, we fixed the object layout the same as show in Fig. \ref{fig-appendix-testset}.

\begin{figure}[t]
\begin{center}
    \includegraphics[width=\linewidth, page=1]{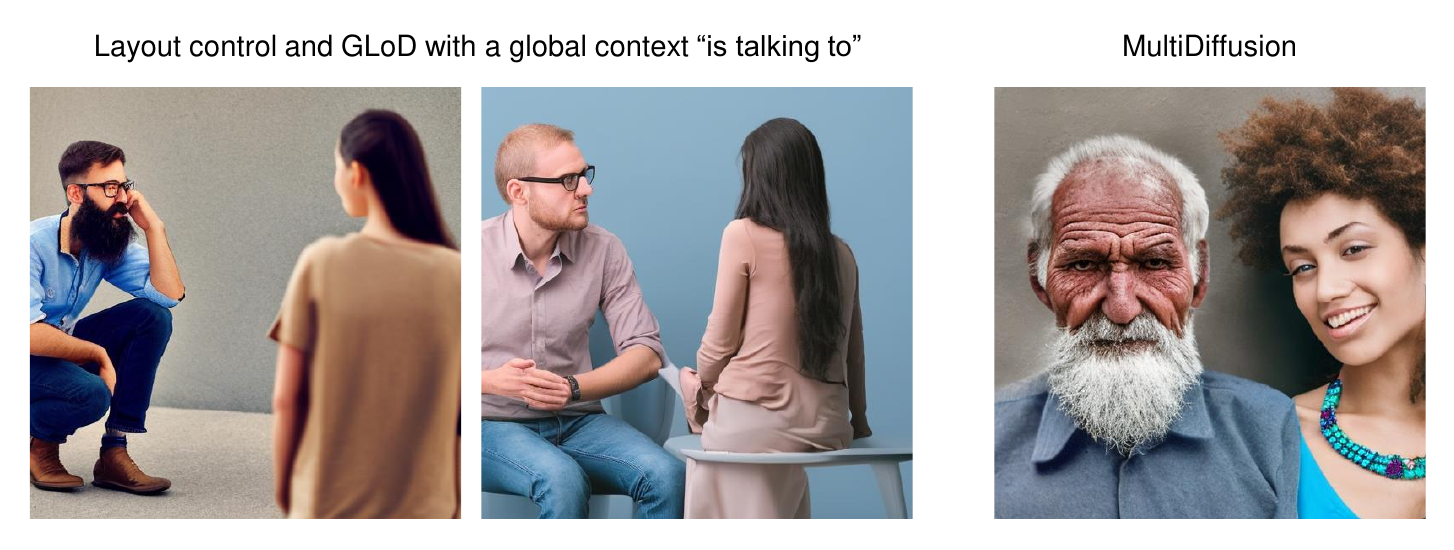}
\end{center}
\caption{Some examples where an object is turning their backs on as a result of complying with the given global context (Left). This does not happen with MultiDiffusion since it does not consider the global context (Right).}
\label{fig-appendix-back}
\end{figure}

\begin{figure}[!t]
\begin{center}
    \includegraphics[width=\linewidth, page=1]{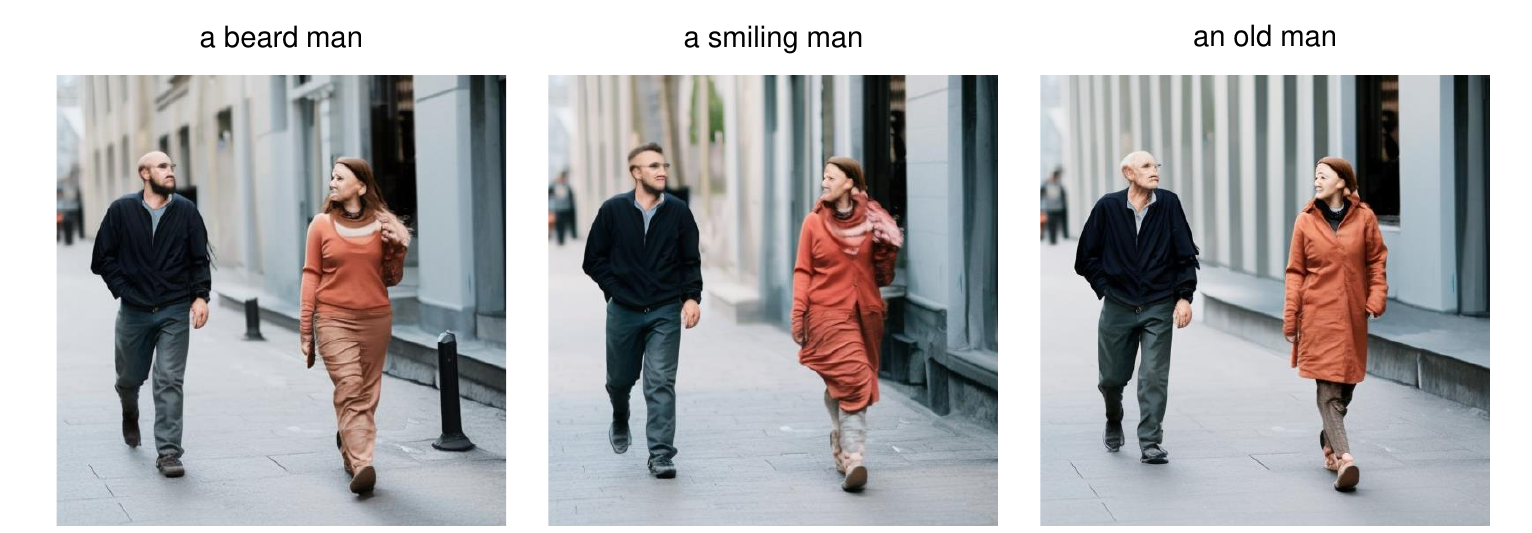}
\end{center}
\caption{GLoD controls the local detail while keeping the global context and the unspecified identities. The visual attribute of the man changes depending on the local text, but the object layout and his posture are almost the same. }
\label{fig-appendix-local}
\end{figure}

\section{Generated samples}
Figure \ref{fig-appendix-back} shows some examples where the alignment score for the object significantly drops. As shown on the left, the subject (i.e., woman) in the generated images are naturally turning their back on since the models try to comply with the given global context ("is talking to"). On the other hand, this does not happen with MultiDiffusion (right) since it does not handle the global context. Therefore, MultiDiffusion achieves the higher alignment score for the objects in Table \ref{tab:scores}.

Figure \ref{fig-appendix-local} shows some examples where GLoD controls the local details. In those images, only the local text for the subject (i.e., man) changes while keeping the global text ("is walking with") and the local text for the object (i.e., woman) the same. As shown in Fig. \ref{fig-appendix-local}, the visual attribute of the man changes depending on the local text, but the object layout and his posture are almost preserved. 

\end{document}